\pdfoutput=1

\documentclass[11pt]{article}

\usepackage{naacl2021}

\usepackage{times}
\usepackage{latexsym}

\usepackage[T1]{fontenc}

\usepackage[utf8]{inputenc}

\usepackage{microtype}

\usepackage{mathtools}

\usepackage{xspace}
\usepackage{enumitem}

\usepackage{graphicx}
\usepackage{amsmath}
\usepackage{amsthm}
\usepackage{booktabs}
\usepackage{algorithm}
\usepackage{algorithmic}
\usepackage{relsize}
\usepackage{adjustbox}
\usepackage{multirow}
\usepackage{arydshln}
\usepackage[font=footnotesize,subrefformat=parens,labelformat=parens]{subfig}
\usepackage{wasysym}

\newcommand{\model}[1]{\text{#1}\xspace}
\newcommand{\mdattn}{\model{C-LSTM}}
\newcommand{\mdcnn}{\model{C-CNN}}
\newcommand{\mdbert}{\model{C-BERT}}

\newcommand{\method}[1]{\textsc{#1}\xspace}

\newcommand{\tsai}{\method{TYC}}
\newcommand{\hotflip}{\method{HotFlip}}
\newcommand{\textfooler}{\method{TextFooler}}

\newcommand{\aevan}{\method{AE}}
\newcommand{\aebal}{\method{AE+bal}}
\newcommand{\aels}{\method{AE+ls}}

\newcommand{\aecf}{\method{AE+ls+cf}}

\newcommand{\aecfcopy}{\method{AE+ls+cf+cpy}}

\newcommand{\dataset}[1]{\texttt{#1}\xspace}
\newcommand{\dsyelpsmall}{\dataset{yelp50}}

\newcommand{\metric}[1]{\text{#1}\xspace}
\newcommand{\accuracy}{\metric{ACC}}
\newcommand{\bleu}{\metric{BLEU}}

\newcommand{\acceptability}{\metric{ACPT}}
\newcommand{\use}{\metric{USE}}
\newcommand{\sent}{\metric{SENT}}
\newcommand{\trans}{\metric{TRF}}

\newcommand{\subsets}[1]{\text{#1}\xspace}
\newcommand{\alls}{\subsets{ALL}}
\newcommand{\poss}{\subsets{POS}}
\newcommand{\negs}{\subsets{NEG}}
\newcommand{\sucs}{\subsets{SUC}}

\newcommand{\agree}{\subsets{AGR}}
\newcommand{\selfcontr}{\subsets{UKN}}
\newcommand{\disagree}{\subsets{DAGR}}

\newcommand{\secref}[2][]{Section#1~\ref{sec:#2}}

\newcommand{\tabref}[2][]{Table#1~\ref{tab:#2}}
\newcommand{\figref}[2][]{Figure#1~\ref{fig:#2}}

\newcommand{\eqnref}[2][]{Equation#1~(\ref{eqn:#2})}

\title{Grey-box Adversarial Attack And Defence For Sentiment Classification}

\author{Ying Xu \thanks{This work was completed during the employment of the authors in IBM Research Australia. }\\
  IBM Research \\Australia \\
\\\And
  Xu Zhong \footnotemark[1] \\
  IBM Research \\ Australia \\
   \\\And
  Antonio Jimeno Yepes \\
  IBM Research \\ Australia \\
    \\\And
  Jey Han Lau \\
  University of Melbourne \\ Australia \\
   \\}

\begin{document}
\maketitle
\begin{abstract}
We introduce a grey-box adversarial attack and defence framework for 
sentiment classification. We address the issues of differentiability, label 
preservation and 
input reconstruction for adversarial attack \textit{and} defence in one 
unified framework.  Our results show that once trained, the attacking 
model is capable of generating high-quality adversarial examples 
substantially faster (one order of magnitude less in time) than 
state-of-the-art attacking methods. These examples also preserve the 
original sentiment according to human evaluation.  Additionally, our 
framework produces an improved classifier that is robust in defending 
against multiple adversarial attacking methods.  Code is available at: 
\url{https://github.com/ibm-aur-nlp/adv-def-text-dist}
\end{abstract}

\section{Introduction}

Recent advances in deep neural networks have created applications for a 
range of different domains.
In spite of the promising performance achieved by neural models, there 
are concerns
around their robustness, as evidence shows that even a slight 
perturbation to the input data can fool these models into producing 
wrong predictions \cite{Goodfellow+:2014,kurakin2016adversarial}.  
Research in this area is broadly categorised as \textit{adversarial 
machine learning}, and it has two sub-fields: \textit{adversarial 
attack}, which seeks to generate adversarial examples that fool target 
models; and \textit{adversarial defence}, whose goal is to build models 
that are less susceptible to adversarial attacks.

A number of adversarial attacking methods have been proposed for image 
recognition \cite{Goodfellow+:2014},  NLP 
\cite{zhang2020adversarial} and speech recognition 
\cite{alzantot2018did}.
These methods are generally categorised into three types: white-box, 
black-box and grey-box attacks. White-box attacks assume full access to 
the target models and
often use the gradients from the target models to guide the craft 
of adversarial examples. 
Black-box attacks, on the other hand, assume no knowledge 
on the architecture of the target model and perform attacks by repetitively 
querying the target model. Different
from the previous two, grey-box attacks train a \textit{generative 
model} to generate adversarial examples and only assume access 
to the target model during the training phrase.
The advantages of grey-box attacking methods include higher time 
efficiency; no assumption of access to target model during attacking phase; and 
easier integration into adversarial defending algorithms. However, due 
to the discrete nature of texts, designing grey-box attacks on text data 
remains a challenge.

In this paper, we propose a grey-box framework that generates 
high quality textual adversarial examples while simultaneously trains an 
improved sentiment classifier for adversarial defending.  Our 
contributions are summarised as follows:

\begin{itemize}[topsep=0pt]
\setlength\itemsep{-0.3em}
\item We propose to use Gumbel-softmax \cite{jang2016categorical} to 
address the differentiability issue to combine the adversarial
example generator and target model into one unified trainable network. 

\item We propose multiple competing objectives for adversarial attack 
training so that the generated adversarial examples can fool the target 
classifier while maintaining similarity with the input examples. 
We considered a number of similarity measures 
to define a successful attacking example for texts, such 
as lexical and semantic similarity and label 
preservation.\footnote{Without constraint on label preservation, simply 
flipping the ground-truth sentiment (e.g.\ \textit{the movie is great} 
$\rightarrow$ \textit{the movie is awful}) can successfully change the 
output of a sentiment classifier even though it is not a useful 
adversarial example. } 

\item To help the generative model to reconstruct input sentences as 
faithfully as possible, we introduce a novel but simple copy mechanism 
to the decoder to selectively copy words directly from the input.

\item We assess the adversarial examples beyond just attacking 
performance, but also content similarity, fluency and label preservation 
using both automatic and human evaluations. 

\item We simultaneously build an improved sentiment classifier while 
training the generative (attacking) model. We show that a classifier 
built this way is more robust than adversarial defending based on 
adversarial examples
augmentation.

\end{itemize}

\section{Related Work}
\label{sec:related}

Most white-box methods are gradient-based, where some form
 of the gradients (e.g.\ the sign) with respect to
the target model is calculated and added to the input 
representation.
In image processing, 
the fast gradient sign method (FGSM; \newcite{Goodfellow+:2014}) is one of 
the first studies in attacking image classifiers. Some of its variations 
include \newcite{kurakin2016adversarial,dong2018boosting}.
These gradient-based methods could not be applied to texts directly
because perturbed word embeddings do not necessarily map to valid
words. Methods such as DeepFool \cite{moosavi2016deepfool}
that rely on perturbing the word embedding space face similar roadblocks.

To address the issue of embedding-to-word mapping, \newcite{gong2018adversarial}
propose to use nearest-neighbour search to find the closest words to the perturbed
embeddings. However, this method treats all tokens as equally vulnerable
and replace all tokens with their nearest neighbours, which leads to 
non-sensical, word-salad outputs. A solution to this
is to replace tokens one-by-one in order of their vulnerability while
monitoring the change of the output of the target models. The replacement
process stops once the target prediction has changed, minimising the 
number of changes. Examples of white-box attacks that utilise this approach 
include \tsai  \cite{tsai2019adversarial} and \hotflip \cite{ebrahimi2017hotflip}.

Different to white-box attacks, black-box attacks do not require full
access to the architecture of the target model. \newcite{chen2017zoo} propose to
estimate the loss function of the target model by querying its 
\textit{label probability distributions}, while 
\newcite{papernot2017practical} propose to construct a substitute of
the target model by querying its \textit{output labels}.  
The latter approach is arguably more realistic 
because in most cases attackers only have access to output labels 
rather than their probability distributions. 
There is relatively fewer studies on black-box attacks for text. 
An example is \textfooler, proposed by \newcite{jin2019bert}, that generates 
adversarial examples by querying the label probability distribution of the target
model.  
Another is proposed by \newcite{alzantot2018generating} where genetic algorithm 
is used to select the word for substitution. 

Grey-box attacks require an additional training process during which 
full access to the target model is assumed. However, post-training, the 
model can be used to generate adversarial examples without querying 
the target model.
\newcite{xiao2018generating} introduce a generative adversarial network
to generate the image perturbation from a noise map. It is, however, not 
trivial to adapt the method for text directly. It is because
text generation involves discrete decoding 
steps and as such the joint generator and target model architecture is 
non-differentiable.

In terms of adversarial defending, the most straightforward method is to
train a robust model on data augmented by adversarial examples. 
Recently, more methods are proposed for texts, such 
as those based on interval bound propagation \cite{jia2019certified,huang2019achieving}, 
and dirichlet neighborhood ensemble \cite{zhou2020defense}. 
%
%

\section{Methodology}

The purpose of adversarial attack is to slightly perturb an input 
example $x$ for a
pre-trained target model (e.g.\ a sentiment classifier) $f$ so that 
$f(x) \neq y$, where $y$ is the ground truth of $x$.
The perturbed example $x'$ should \textit{look} similar to $x$,
which can be measured differently depending on the domain of 
the input examples.

\subsection{General Architecture}
\label{sec:general-architecture}


We propose a grey-box attack and defence framework which consists of a 
generator $\mathcal{G}$ (updated), and two copies of a pre-trained target 
classifier: a static classifier $\mathcal{C}$ and an updated/augmented 
classifier $\mathcal{C^{*}}$.\footnote{$\mathcal{C}$ and 
$\mathcal{C^{*}}$ start with the same pre-trained weights, although only 
$\mathcal{C^{*}}$ is updated during training.}
During the training phase, the output of $\mathcal{G}$ is directly fed 
to $\mathcal{C}$ and $\mathcal{C^{*}}$ to form a joint architecture.  
Post-training, the generator $\mathcal{G}$ is used independently to 
generate adversarial examples (adversarial attack);  while the augmented
classifier $\mathcal{C^*}$ is an improved classifier with increased 
robustness (adversarial defence).

The training phase is divided into attacking steps and defending steps, 
where the former updates only the generator $\mathcal{G}$ and learns to 
introduce slight perturbation
to the input by maximising the objective function of the target 
model $\mathcal{C}$. The latter updates $\mathcal{C^{*}}$ and $\mathcal{G}$ by 
feeding both original examples and adversarial examples 
generated by $\mathcal{G}$. 
Here, the adversarial examples are assumed to share the same label with their 
original examples. 
Effectively, the defending steps are training an improved classifier with data 
augmented by adversarial examples.

\begin{figure}[t]
\centering
\includegraphics[width=0.4\textwidth]{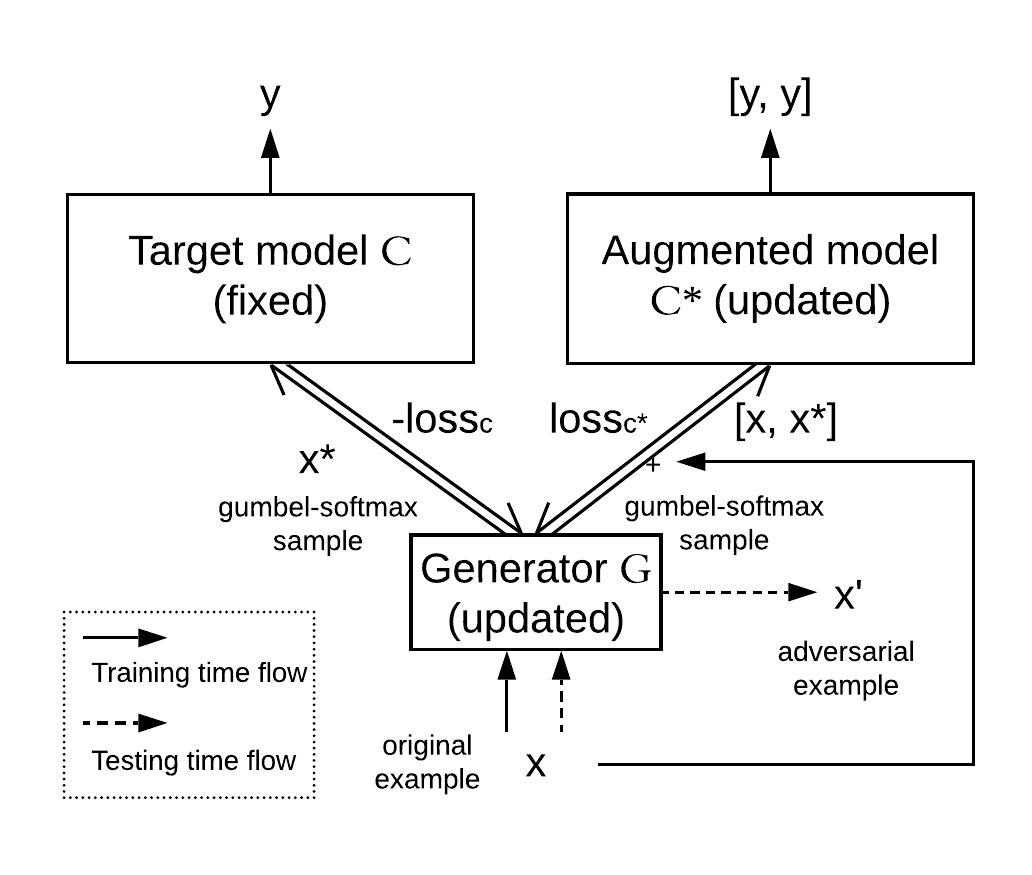}
  \caption{Grey-box adversarial attack and defence framework for sentiment classification. }
  \label{fig:framework}
\end{figure}

Generating text with discrete decoding steps (e.g. \textit{argmax}) makes 
the joint architecture not differentiable.
Therefore we propose to use Gumbel-softmax \cite{jang2016categorical} 
to approximate the categorical distribution of the discrete output. 
For each generation step $i$, instead of sampling a word from the vocabulary, 
we draw a Gumbel-softmax sample $x_i^{*}$ which has the full probability 
distribution over words in the vocabulary: the probability of the generated word is 
close to 1.0 and other words close to zero.
We obtain the input embedding for $\mathcal{C}$ and 
$\mathcal{C^{*}}$ by multiplying the sample $x_i^*$ with the word 
embedding matrix, $M_{\mathcal{C}}$, of the target model $\mathcal{C}$: 
$x_i^{*} \cdot M_{\mathcal{C}}$.
\figref{framework} illustrates our grey-box adversarial 
attack and defence framework for text.

The generator $\mathcal{G}$ can be implemented as an auto-encoder
or a paraphrase generator, essentially differentiated by their data 
conditions: the former uses the input sentences as the target, while the 
latter uses paraphrases (e.g.\ PARANMT-50M \cite{wieting2017paranmt}).
In this paper, we implement $\mathcal{G}$ as an auto-encoder, as our preliminary 
experiments found that a pre-trained paraphrase generator performs 
poorly when adapted to our test domain, e.g. Yelp reviews. 

\subsection{Objective Functions}
\label{sec:objective-functions}

Our auto-encoder $\mathcal{G}$ generates an adversarial example given an input 
example.
It tries to reconstruct the input example but is also regulated by an adversarial 
loss term that ‘discourages’ it from doing so. 
The objectives for the attacking step are given as follows:
\begin{align}
L_{adv} &= \log p_{\mathcal{C}}(y|x, \theta_{\mathcal{C}}, \theta_{\mathcal{G}})
\label{eqn:adv}\\
L_{s2s} &= -\log p_{\mathcal{G}} (x | x, \theta_{\mathcal{G}})\\
L_{sem} &= cos \left (\frac{1}{n}\sum_{i=0}^{n}emb(x_i), 
\frac{1}{n}\sum_{i=0}^{n}emb(x^*_i)\right )
\end{align}

\noindent where $L_{adv}$ is essentially the negative cross-entropy loss 
of $\mathcal{C}$; $L_{s2s}$ is the sequence-to-sequence loss for input 
reconstruction;
and $L_{sem}$ is the cosine similarity between the averaged embeddings 
of $x$ and $x^*$ ($n =$ number of words).  Here, $L_{s2s}$ encourages 
$x'$ (produced at test time) to be lexically similar to $x$ and helps 
produce coherent sentences, and $L_{sem}$ promotes semantic similarity.  
We weigh the three objective
 functions with two scaling hyper-parameters and the total loss is: 
$L = \lambda_{1}(\lambda_{2} L_{s2s} + (1-\lambda_{2})L_{sem}) + (1- \lambda_{1})L_{adv}$
We denote the auto-encoder based generator trained with these objectives 
as \textbf{\aevan}.

An observation from our preliminary experiments is that the generator 
tends to perform imbalanced attacking among different classes. 
 (e.g.\ \aevan learns to completely focus on one direction attacking, 
e.g.\
 positive-to-negative or negative-to-positive attack).
We found a similar issue in white-box attack methods such as FGSM \newcite{Goodfellow+:2014} 
and DeepFool \cite{moosavi2016deepfool}. 
To address
this issue, we propose to modify $L_{adv}$ to be the maximum
loss of a particular class in each batch, i.e.\  

\begin{equation}
L_{adv} = max_{t=1}^{|C|}(L_{adv}^{t})
\end{equation}

where $L_{adv}^{t}$ refers to the adversarial loss of examples in the $t$-th class and
 $|C|$ the total number of classes. We denote the generator 
trained with this alternative loss as \textbf{\aebal}.


For adversarial defence, we use the same objective functions, with the 
following exception: we replace $L_{adv}$
in \eqnref{adv} with the objective function of the classifier 
$\mathcal{C^{*}}$, i.e.

\begin{equation}
L_{def}=-\log p_{\mathcal{C^*}}([y, y]|[x, x^*], \theta_{\mathcal{C^*}}, \theta_{\mathcal{G}})
\end{equation}

We train  the model $\mathcal{C^*}$ using both original and adversarial 
examples ($x$ and $x^*$) with their original label ($y$) to prevent 
$\mathcal{C^*}$ from overfitting to the adversarial examples.  

\subsection{Label Preservation}
\label{sec:label-preservation}

One of the main challenges of generating a textual adversarial example 
is to preserve its original ground truth label, which we refer to as 
\textit{label preservation}. It is less of an issue in computer vision 
because slight noises added to an image is unlikely to change 
how we perceive the image. In text, however, slight perturbation to 
a sentence could completely change its ground truth. 

We use sentiment classification as context to explain our approach for 
label preservation.
The goal of adversarial attack is to
generate an adversarial sentence whose sentiment is flipped according 
to the target model prediction
but preserves the original ground truth sentiment from the perspective 
of a human reader. 
We propose two ways to help label preservation.  The first approach is 
task-agnostic, i.e.\ it can work for any classification problem, while 
the second is tailored for sentiment classification.


\textbf{Label smoothing (\method{\textbf{+ls}})}. We observe the generator has a tendency to 
produce adversarial examples with high confidence, opposite sentiment 
scores from the static classifier $\mathcal{C}$. We explore the use of 
label smoothing \cite{muller2019does} to force the generator generate 
examples that are closer to the decision boundary, to discourage the 
generator from completely changing the sentiment. We incorporate label smoothing
in Eq. \ref{eqn:adv} by redistributing the probability mass of true label uniformly to all 
other labels.
Formally, the smoothed label $y_{ls} = (1 - \alpha) * y + \alpha / K$ where 
$\alpha$ is a hyper-parameter and $K$ is the number of classes.
For example, when performing negative-to-positive attack, instead of optimising
$\mathcal{G}$ to produce adversarial examples with label distribution \{pos: 1.0, neg: 0.0\} 
(from $\mathcal{C}$), label distribution \{pos: 0.6, neg: 0.4\} is targeted.  
Generator trained with this additional constraint is denoted with the \method{\textbf{+ls}} suffix.

\textbf{Counter-fitted embeddings (\method{\textbf{+cf}})}. \newcite{mrkvsic2016counter} found 
that unsupervised word embeddings such as GloVe \cite{pennington2014glove}
often do not capture synonymy and antonymy relations (e.g.\ 
\textit{cheap} and \textit{pricey} have high similarity).
The authors propose to post-process pre-trained word embeddings with 
lexical resources (e.g.\ WordNet) to produce counter-fitted embeddings 
that better capture these lexical relations.
To discourage the generator $\mathcal{G}$ from generating words with 
opposite sentiments, we experiment with training $\mathcal{G}$ with 
counter-fitted embeddings. Models using counter-fitted embeddings is 
denoted with \method{\textbf{+cf}} suffix.


\subsection{Generator with Copy Mechanism (\method{\textbf{+cpy}})} 
\label{sec:copy}



White-box or black-box attacking methods are based on adding, removing,
or replacing tokens in input examples. Therefore maintaining similarity with 
original examples is easier than grey-box methods
that generate adversarial examples word-by-word from scratch. We introduce 
a simple copy mechanism that helps grey-box attack to produce faithful 
reconstruction of the original sentences. 

We incorporate a static copy mask to the 
decoder where it only generates for word positions that have not been  
masked.  E.g., given the input sentence $x = [w_0, w_1, w_2]$, target $x^* 
= [w_0, w_1, w_2]$, and mask $m = [1, 0, 1]$, at test time the decoder 
will ``copy'' from the target for the first input ($w_0$) and third 
input token ($w_2$) to produce $w_0$ and $w_2$, but for the second input 
token ($w_1$) it will decode from the vocabulary. During training, we 
compute cross-entropy only for the unmasked input words.

The static copy mask is obtained from one of the pre-trained target 
classifiers, \mdattn (\secref{implementation-details}). \mdattn is a 
classifier with a bidirectional LSTM followed by a self-attention layer 
to weigh the LSTM hidden states. We rank the input words based on the 
self-attention weights and create a copy mask such that only the 
positions corresponding to the top-$N$ words with the highest weights 
are generated from the decoder. Generally sentiment-heavy words such as 
\textit{awesome} and \textit{bad} are more likely to have higher weights 
in the self-attention layer.  This self attention layer can be seen as an 
importance ranking function \cite{morris2020textattack} that determines 
which tokens should be replaced or replaced first. 
Models with copy mechanism are denoted 
with the \method{\textbf{+cpy}} suffix.

\section{Experiments and Results}

\subsection{Dataset}

We conduct our experiments using the Yelp review 
dataset.\footnote{\url{https://www.yelp.com/dataset}} We binarise the 
ratings,\footnote{Ratings$\geq$4 is set as positive and ratings$\leq$2
as negative.} use spaCy for 
tokenisation,\footnote{\url{https://spacy.io}} and keep only reviews 
with $\leq$ 50 tokens (hence the dataset is denoted as \dsyelpsmall).  
We split the data in a 90/5/5 ratio and downsample the positive class in 
each set to be equivalent to the negative class, resulting in 407,298, 
22,536 and 22,608 examples in train/dev/test set respectively. 

\subsection{Implementation Details}
\label{sec:implementation-details}

For the target classifiers ($\mathcal{C}$ and $\mathcal{C^*}$),
we pre-train three sentiment classification models using \dsyelpsmall: 
\mdattn \cite{wang2016attention},  \mdcnn \cite{kim2014convolutional} and \mdbert.
\mdattn is composed of an embedding layer, a 2-layer bidirectional 
LSTMs, a self-attention layer, and an output layer.
\mdcnn has a number of convolutional filters of
varying sizes, and their outputs are concatenated, pooled and fed to
a fully-connected layer followed by an output layer.
Finally, \mdbert is obtained by fine-tuning the BERT-Base model
\cite{devlin2018bert} for sentiment classification. We tune learning 
rate, batch size, number of layers and number of hidden units for all 
classifiers; the number of attention units for \mdattn and 
convolutional filter sizes and dropout rates for \mdcnn specifically. 

For the auto-encoder, we pre-train it to reconstruct sentences in 
\dsyelpsmall.\footnote{Pre-trained \bleu scores are 97.7 and 96.8 on 
\dsyelpsmall using GloVe and counter-fitted embedding, respectively. }
During pre-training, we tune learning rate, batch size, number of layers and 
number of hidden units. During the training of adversarial attacking, we tune
$\lambda_{1}$ and $\lambda_{2}$, and learning rate $lr$.
We also test different temperature $\tau$ for Gumbel-softmax sampling
 and found that $\tau=0.1$ performs the best.  
All word embeddings are fixed.

More hyper-parameter and training configurations are detailed in the 
supplementary material.

\subsection{Attacking Performance}
\label{sec:attacking-performance}

Most of the existing adversarial attacking methods have been focusing on 
improving the attack success rate. Recent study show that with constraints 
adjusted to better preserve semantics and grammaticality, the attack success 
rate drops by over 70 percentage points \cite{morris2020reevaluating}. 
In this paper, we want to understand --- given a particular success rate --- 
the quality (e.g. fluency, content/label preservation) of the generated adversarial 
samples. Therefore, we tuned all attacking methods to achieve 
the same levels of attack success rates; and compare the quality of generated 
adversarial examples. 
\footnote{We can in theory tune different methods to achieve higher success rate, 
but we choose the strategy to use lower success rates so that all methods generate 
relatively fair quality examples that annotators can make sense of during human 
evaluation. 
} Note that results for adversarial attack are 
obtained by using the $\mathcal{G}+\mathcal{C}$ joint architecture,
while results for adversarial defence are achieved by the 
$\mathcal{G}+\mathcal{C}+\mathcal{C^*}$ joint architecture.


\subsubsection{Evaluation Metrics}

In addition to measuring how well the adversarial examples fool the 
sentiment classifier, we also use a number of automatic metrics to 
assess other aspects of adversarial examples, following 
\newcite{xu2020elephant}:

\textbf{Attacking performance}. We use the standard classification 
accuracy (\accuracy) of the target classifier ($\mathcal{C}$)
to measure the attacking performance of adversarial examples. 
Lower accuracy means better attacking performance.

\textbf{Similarity}. To assess the textual and semantic similarity between the 
original and corresponding adversarial examples, we compute \bleu 
\cite{Papineni+:2002} and \use \cite{cer2018universal}.\footnote{\use is 
calculated as the cosine similarity between the original and 
adversarial sentence embeddings produced by the universal sentence 
encoder \cite{cer2018universal}.} For both metrics, higher scores 
represent better performance.

\textbf{Fluency}. To measure the readability of generated adversarial examples,
we use the acceptability score (\acceptability) proposed by 
\newcite{Lau+:2020}, which is based
on normalised sentence probabilities produced by XLNet 
\cite{Yang+:2019}. Higher scores indicate better fluency.

\textbf{Transferability}. To understand the effectiveness of the 
adversarial examples in attacking another unseen sentiment classifier 
(\trans), we evaluate the accuracy
of \mdbert using adversarial examples that have been generated for 
attacking classifiers \mdattn and \mdcnn. Lower accuracy indicates 
better transferability.

\textbf{Attacking speed}. We measure each attacking method on the amount 
of time  it takes on average (in seconds) to generate an adversarial 
example.


\subsubsection{Automatic Evaluation}

\textbf{Comparison between \aevan variants.} We first present results on 
the \textit{development set} where we explore different variants of the 
auto-encoder (generator) in the grey-box model.  \aevan serves as our 
base model, the suffix \method{+bal} denotes the use of an alternative 
$L_{adv}$ (\secref{objective-functions}),  \method{+ls}  label smoothing 
(\secref{label-preservation}), \method{+cf} counter-fitted embeddings 
(\secref{label-preservation}), and \method{+cpy} copy mechanism 
(\secref{copy}).

We present the results in \tabref{variations}. Attacking performance of 
all variants are tuned to produce approximately 70\% -- 80\% accuracy for the 
target classifier $\mathcal{C}$ (\mdattn).
For \accuracy and \bleu, we additionally report the performance for the 
positive and negative sentiment class separately. To understand how well 
the adversarial examples preserve the original sentiments,  we recruit 
two annotators internally to annotate a small sample of adversarial 
examples
produced by each of the auto-encoder variants. \agree and \disagree 
indicate the percentage of adversarial examples where they agree and 
disagree with the original sentiments, and \selfcontr where the 
annotators are unable to judge their sentiments.

\begin{table}
\begin{center}
\begin{adjustbox}{max width=\linewidth}
\begin{tabular}{lc@{\;\;}c@{\;\;}cc@{\;\;}c@{\;\;}cc@{\;\;}c@{\;\;}c}
\toprule
  & \multicolumn{3}{c}{\accuracy} & \multicolumn{3}{c}{\bleu} & \multicolumn{3}{c}{\sent} \\
  & \alls & \poss & \negs & \sucs & \poss & \negs & \agree & \selfcontr & \disagree \\\midrule
 \aevan & 66.0 & 99.8 & 28.4 & 55.3 & 71.7 & 58.7 & -- &  --   & --  \\
  \aebal & 75.6 & 72.3 & 78.8 & 65.9 & 73.9 & 70.9 & 0.12 & 0.80 & 0.08 \\
  \aels & 74.3 & 77.8 & 70.4 & 80.3 & 84.6 & 86.3 & 0.46 & 0.44 & 0.10 \\
 \aecf & 76.6 & 66.5 & 86.7 & 79.9 & 82.5 & 85.0 & 0.64 & 0.28 & 0.08 \\
 \aecfcopy & 77.4 & 70.9 & 83.8 & \textbf{85.7} & 90.6 & 90.2 & \textbf{0.68} & 0.30 & 0.02  \\
\bottomrule
\end{tabular}
\end{adjustbox}
\end{center}
\caption{Performance of adversarial examples generated by five 
\aevan variants on \dsyelpsmall development set.  }
\label{tab:variations}
\end{table}

\begin{table*}[t!]
\begin{center}
\small
\begin{adjustbox}{max width=0.6\linewidth}
\begin{tabular}{llc@{\;}c@{\;}c@{\;}c@{\;}cc@{\;}c@{\;}c@{\;}c@{\;}c}
\toprule
&& \multicolumn{5}{c}{\mdattn: 96.8}  &
 \multicolumn{5}{c}{\mdcnn: 94.3} \\
& {Model} & \accuracy & \bleu & \use & \acceptability &\trans&
  \accuracy & \bleu & \use & \acceptability &\trans \\
 \midrule
\multirow{4}{*}{T1}
 & \tsai & 83.8 & 48.3 &  11.6 & -18.9 & \textbf{87.1} & 87.6 & 41.2 & 29.4 & -21.8 & \textbf{91.5} \\
 & \hotflip & 80.3 & 85.6 & 47.9 & -7.0 & 93.3 & 81.5 & \textbf{92.5} & 77.1 & -3.8 & 95.1 \\
   & \textfooler & 86.5 & \textbf{92.6} & \textbf{88.7} & \textbf{-1.8} & 94.6 & 87.7 & 91.9 & \textbf{94.2} & \textbf{-2.1}& 96.2  \\
 & \aecfcopy & 87.7 & 86.8 & 83.5 & -3.8 & 95.0 & 85.1 & 87.8 & 80.7 & -4.1 & 94.8 \\
 \midrule
\multirow{4}{*}{T2}
 & \tsai & 75.3 & 41.2 & -7.6 & -20.7 & \textbf{78.2} & 73.4 & 38.9 & -15.3 & -21.4 & \textbf{75.9} \\
 & \hotflip & 75.3 & 80.0 & 38.1  & -7.8 & 91.7 & 70.8 & \textbf{84.7} & 63.4 & -7.1 & 93.4 \\
 & \textfooler & 73.6 & \textbf{88.5} & \textbf{84.1} & \textbf{-2.9} & 92.8 & -- & -- & -- & -- & --  \\
 & \aecfcopy & 77.1 & 83.5 & 74.6 & -5.6 & 92.6 & 78.3 & 82.6 & \textbf{70.2} & \textbf{-5.7} & 92.5 \\
 \midrule
\multirow{4}{*}{T3}
 & \tsai &  65.5 & 30.1  & -7.3 & -26.4 & \textbf{68.7} & -- & -- & -- & -- & --  \\
 & \hotflip & 62.5 & 77.7  & 36.3 & -9.9 & 91.2 & 67.1 & \textbf{81.8} & 57.2 & -8.0 & 92.8  \\
 & \textfooler & 62.2 & \textbf{85.6}  & \textbf{82.6} & \textbf{-3.7} & 91.7 & -- & -- & -- & -- & --   \\
 & \aecfcopy & 66.5 & 80.2 & 67.0 & -7.3 & 90.1 & 69.1 & 76.4 & \textbf{61.7} & \textbf{-7.7} & \textbf{91.7}  \\
\bottomrule
\end{tabular}
\end{adjustbox}
\end{center}
\caption{Results based on automatic metrics, with \mdattn and \mdcnn as 
target classifiers. Dashed line indicates the model is unable to 
generate adversarial examples that meet the accuracy threshold. The 
numbers next to the classifiers (\mdattn and \mdcnn) are the pre-trained 
classification accuracy performance.
}
\label{tab:rc-full}
\end{table*}

\begin{table}[t]
\scriptsize
\begin{center}
\begin{tabular}{p{0.13\textwidth}p{0.32\textwidth}}
\toprule
Direction & neg-to-pos   \\ \midrule
Original & unresonable and hard to deal with ! avoid when looking into a home . plenty of headaches . \\ \midrule
\tsai & \textcolor{red}{homeschoolers} and \textcolor{red}{tantrumming} to \textcolor{red}{marker} with ! \textcolor{red}{australasia blerg quotation} into a home . plenty of headaches . \\
\hotflip & unresonable and \textcolor{red}{ginko} to deal with ! avoid when looking into a home . plenty of headaches .  \\
\textfooler & unresonable and \textcolor{red}{tough} to deal with ! \textcolor{red}{avoids} when looking into a home . plenty of headaches .  \\
\aecfcopy & unresonable and hard to deal with ! \textcolor{red}{canceling} when looking into a home . plenty of headaches . \\
\midrule
\midrule
Direction & pos-to-neg   \\ \midrule
Original & i wish more business operated like this . these guys were all awesome . very organized and pro \\ \midrule
\tsai & \textcolor{red}{relly tthe smushes gazebos slobbering americanised expiration 3.88 magan colered 100/5 bellevue destine 3.88} very \textcolor{red}{02/16 wonderfuly whelms} \\
\hotflip & i wish more business operated \textcolor{red}{a} this . these guys \textcolor{red}{cpp} all \textcolor{red}{stereotypic} . very \textcolor{red}{provisioned} and pro  \\
\textfooler & i wish more business operated \textcolor{red}{iike} this . these guys were all \textcolor{red}{magnificent} . very organized and pro  \\
\aecfcopy & i wish more business operated like this . these guys were all \textcolor{red}{impresses} . very organized and pro \\
\bottomrule
\end{tabular}
\end{center}
\caption{Adversarial examples generated by different methods when attacking on \dsyelpsmall at threshold T2. }
\label{tab:rc-examples}
\end{table}

Looking at the ``POS'' and ``NEG'' performance of \aevan and \aebal, we 
can see that \aebal is effective in creating a more balanced performance 
for positive-to-negative and negative-to-positive attacks.  We 
hypothesise that \aevan learns to perform single direction attack 
because it is easier to generate positive (or negative) words for all 
input examples and sacrifice performance in the other direction to 
achieve a particular attacking performance. That said, the low \agree 
score (0.12) suggests that \aebal adversarial examples do not preserve 
the ground truth sentiments.

The introduction of label smoothing (\aels) and counter-fitted 
embeddings (\aecf) appear to address label preservation, as \agree 
improves from 0.12 to 0.46 to 0.64.
Adding the copy mechanism (\aecfcopy) provides also some marginal 
improvement, although the more significant benefit is in sentence 
reconstruction: a boost of 5 \bleu points.  Note that we also 
experimented with incorporating \method{+bal} for these variants, but 
found minimal benefit.
For the rest of the experiments, we use \aecfcopy as our model to 
benchmark against other adversarial methods.


\textbf{Comparison with baselines.} We next present results on the \textit{test set} in \tabref{rc-full}.  
The benchmark methods are: \tsai, \hotflip, and \textfooler (described 
in \secref{related}). We choose 3 \accuracy thresholds as the basis for 
comparison: T1, T2 and T3, which correspond to approximately 80-90\%, 
70-80\% and 60-70\% accuracy.\footnote{We tune hyper-parameters for each 
attacking method to achieve the 3 attacking thresholds.  }


Generally, all models trade off example quality for attacking rate, 
as indicated by the lower \bleu, \use and \acceptability scores at
T3.

Comparing \mdattn and \mdcnn, we found that \mdcnn is generally an
easier classifier to attack, as  \bleu and \use scores for the same 
threshold are higher. Interestingly, \textfooler appears to be 
 ineffective for attacking \mdcnn, as we are unable to tune 
\textfooler to generate adversarial examples producing \accuracy 
below the T1 threshold.

Comparing the attacking models and focusing on \mdattn, \textfooler 
generally has the upper hand. \aecfcopy performs relatively well, and 
usually not far behind \textfooler.
\hotflip produces good \bleu scores, but substantially worse \use 
scores. 
\tsai is the worst performing model, although its adversarial examples 
are good at fooling the unseen classifier \mdbert (lower \trans than all 
other models), suggesting that there may be a (negative) correlation 
between in-domain performance and transferability.
Overall, most methods do not produce adversarial examples that 
are very effective at attacking \mdbert.\footnote{The  sentiment 
classification accuracy for \mdbert on \dsyelpsmall is originally 97.0.}

\textbf{Case study.} In Table \ref{tab:rc-examples}, we present two randomly selected 
adversarial examples (positive-to-negative and negative-to-positive) for 
which
all five attacking methods successfully fool \mdattn.
\tsai produces largely gibberish output.
\hotflip tends to replace words with low semantic similarity with the 
original words (e.g.\ replacing
\textit{hard} with \textit{ginko}), which explains its high \bleu scores 
and
low \use and \acceptability scores.
Both \textfooler and \aecfcopy generate adversarial examples that are 
fluent and generally retain their original meanings.  These 
observations agree with the quantitative performance we see 
in \tabref{rc-full}.

\textbf{Time efficiency.} Lastly, we report the time it takes for these methods to perform 
attacking on \dsyelpsmall at T2. The average time taken per 
example (on GPU v100) are: 1.2s for \tsai; 1s for \textfooler; 0.3s for \hotflip; and 0.03s 
for \aecfcopy.
\tsai and \textfooler are the slowest methods, while \hotflip is 
substantially faster.  Our model \aecfcopy is the fastest method: about 
an order of magnitude faster compared to the next best method \hotflip. 
Though one should be noted that our grey-box method requires an additional step of 
training that can be conducted offline.


\subsubsection{Human Evaluation}
\label{sec:human-evaluation}

Automatic metrics provide a proxy to quantify the quality of the
adversarial examples. To validate that these metrics work,
we conduct a crowdsourcing experiment on 
Appen.\footnote{\url{https://www.appen.com}}

We test the 3 best performing models (\hotflip, \textfooler and
\aecfcopy) on 2 attacking thresholds (T2 and T3). For each method, we randomly
sampled 25 positive-to-negative and 25 negative-to-positive successful 
adversarial examples. For quality control, we annotate 10\% of the 
samples
 as control questions. Workers are first presented with a 10-question 
quiz, and only
those who pass the quiz with at least 80\% accuracy can work on the
task. We monitor work quality throughout the annotation process by 
embedding a quality-control question in every 10 questions, and stop 
workers from continuing on the task whenever their accuracy on the 
control questions fall below 80\%.  We restrict our jobs to
workers in United States, United Kingdom, Australia, and Canada.

We ask crowdworkers the following questions:

\begin{enumerate}[topsep=0pt]
\setlength\itemsep{-0.3em}
  \item Is snippet B a good paraphrase of snippet A? \newline
  \hspace{4pt} $\ocircle$ Yes \hspace{4pt} $\ocircle$ Somewhat yes \hspace{4pt} $\ocircle$ No
  \item How natural does the text read? \newline
  \hspace{4pt} $\ocircle$ Very unnatural \hspace{4pt} $\ocircle$ Somewhat natural \\ \hspace{4pt} $\ocircle$ Natural
   \item What is the sentiment of the text? \newline
  \hspace{4pt} $\ocircle$ Positive \hspace{4pt} $\ocircle$ Negative \hspace{4pt} $\ocircle$ Cannot tell
\end{enumerate}

We display both the original and adversarial examples for question 1, 
and only the adversarial example for question 2 and 3. As a baseline, we 
also select 50 random original sentences from the test set and collect 
human judgements for these sentences on question 
2 and 3.

\begin{figure}[!t]
  \begin{minipage}[b]{0.48\textwidth}%
    \centering
    \subfloat[Original
  examples]{\includegraphics[width=\linewidth]{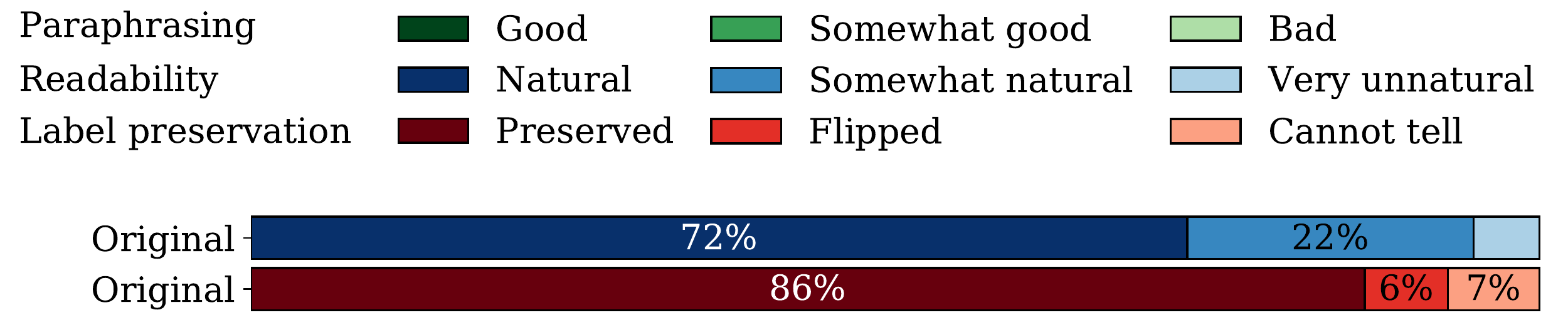}}%
  \end{minipage}\\
  \hfill
  \begin{minipage}[b]{0.48\textwidth}%
    \centering
    \subfloat[\accuracy threshold:
  T2]{\includegraphics[width=\textwidth]{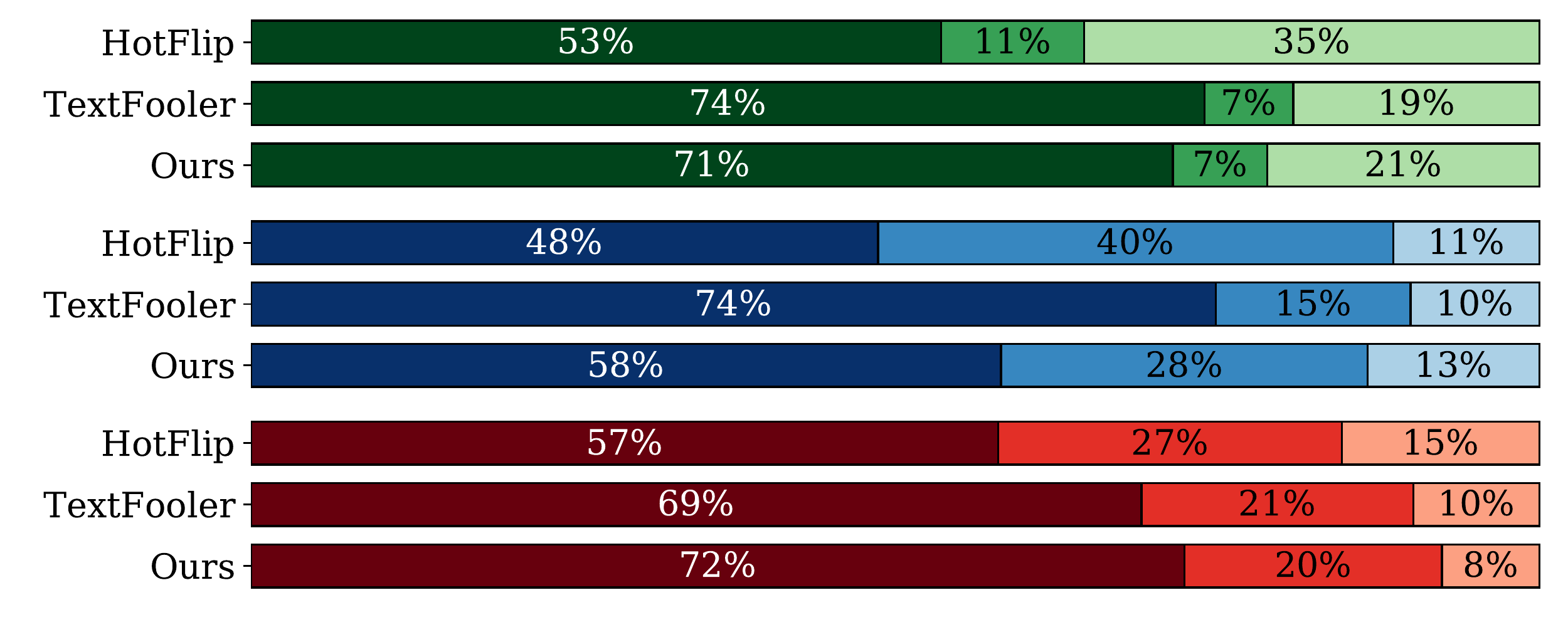}}%
  \end{minipage}\\
  \hfill
  \begin{minipage}[b]{0.48\textwidth}%
    \centering
    \subfloat[\accuracy threshold:
  T3]{\includegraphics[width=\textwidth]{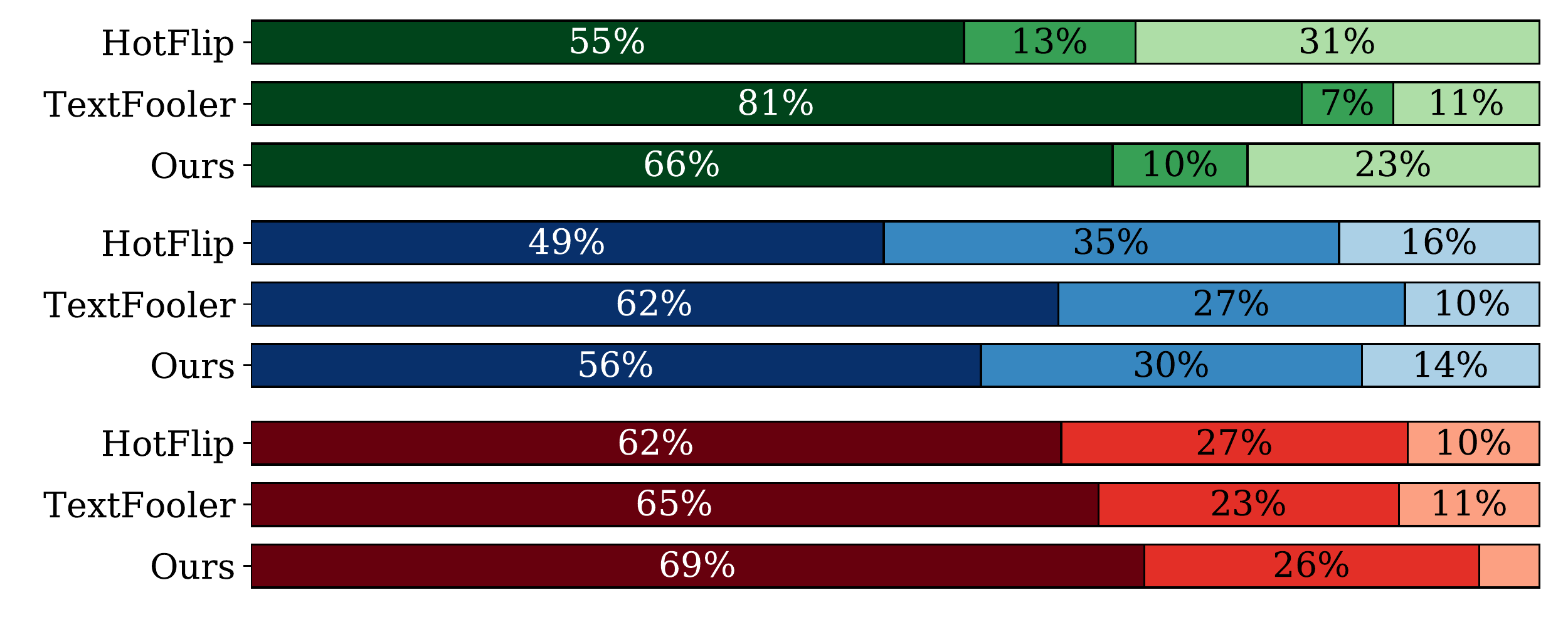}}%
  \end{minipage}
  \caption{Human evaluation results.}
  \label{fig:human_eval}
\end{figure}

We present the human evaluation results in Figure \ref{fig:human_eval}.  
Looking at the original examples (top-2 bars), we see that they are 
fluent and their perceived sentiments (by the crowdworkers) have a high 
agreement with their original sentiments (by the review authors).
Comparing the 3 methods, \textfooler produces adversarial sentences that 
are most similar to the original (green) and they are more natural 
(blue) than other methods. \hotflip is the least impressive method here, 
and these observations agree with the scores of automatic metrics 
in \tabref{rc-full}. On label preservation (red), however, our method 
\aecfcopy has the best performance, implying that the generated 
adversarial sentences largely preserve the original sentiments. 

The consistency between the automatic and human evaluation results 
indicate that the \use and \acceptability scores properly captured the 
semantic similarity and readability, two important evaluation aspects 
that are text-specific.


\subsection{Defending Performance}

Here we look at how well the generated adversarial examples can help 
build a more robust classifier. Unlike the attacking performance 
experiments (\secref{attacking-performance}), here we include the 
augmented classifier ($\mathcal{C^*}$) as part of the grey-box 
training.\footnote{During training, we perform one attacking step for 
every two defending steps.} The augmented classifier can be seen as an 
improved model compared to the original classifier $\mathcal{C}$. 

To validate the performance of adversarial defence, we evaluate the 
accuracy of the augmented classifiers against different attacking 
methods. We compared our augmented classifier $\mathcal{C^*}$ to the 
augmented classifiers adversarially trained with adversarial examples 
generated from \hotflip and \textfooler. Our preliminary results show that 
training $\mathcal{C^*}$ without the copy mechanism provides better defending
performance, therefore we use the \aecf architecture to obtain $\mathcal{C^*}$. 

For fair comparison, our augmented classifier ($\mathcal{C^*}$) is 
obtained by training the generator ($\mathcal{G}$) to produce an 
attacking performance of T2 accuracy (70\%) on the static classifier 
($\mathcal{C}$).  
For the other two methods, we train an
augmented version of the classifier by feeding the
original training data together with the adversarial examples \footnote{
one per each training example} generated 
by \hotflip and \textfooler with the same T2 attacking performance; 
these two classifiers are denoted as $\mathcal{C}_\textfooler$ and 
$\mathcal{C}_\hotflip$, respectively.

At test time, we attack the three augmented classifiers using \tsai, 
\hotflip, \textfooler and
\aecf, 
and evaluate their classification accuracy.
Results are presented in \tabref{rc-defence}.  The second row ``Original 
Perf.'' indicates the performance when we use the original test examples 
as input to the augmented classifiers. We see a high accuracy here, 
indicating that the augmented classifiers still perform well on the 
original data.

%

\begin{table}[t]
\begin{center}
\begin{adjustbox}{max width=0.7\linewidth}
\begin{tabular}{lcccc}
\toprule
 &  $\mathcal{C}$  &  $\mathcal{C}_{\hotflip}$  &  
$\mathcal{C}_{\textfooler}$  &   $\mathcal{C}^{*}$ \\ \midrule
Original Perf.   & 96.8 & 96.6 & 96.9 & \textbf{97.2} \\ \hdashline
\tsai  & 75.3 & 69.5 & 73.1 & \textbf{76.0} \\
\hotflip  & 75.3 & 61.2 & 80.1 &  \textbf{97.1} \\
\textfooler & 73.6 & 66.4 & 74.5 & \textbf{76.5} \\
\aecf  & 74.0 & 83.2 & 86.3 & \textbf{90.0} \\
\bottomrule
\end{tabular}
\end{adjustbox}
\end{center}
\caption{Defending performance.}
\label{tab:rc-defence}
\end{table}

Comparing the different augmented classifiers, our augmented classifier
$\mathcal{C^*}$ outperforms the other two in defending against different 
adversarial attacking methods (it is particularly good against \hotflip).
It produces the largest classification improvement compared to the 
original classifier $\mathcal{C}$ (0.7, 
21.8, 2.9 and 16.0 points against adversarial examples created by \tsai, \hotflip,
\textfooler and \aecf  respectively).
Interestingly, the augmented classifier trained with \hotflip adversarial 
examples ($\mathcal{C}_{\hotflip}$) produces a more vulnerable model, as it has 
lower accuracy compared to original classifier ($\mathcal{C}$).
We suspect this as a result of training with low quality adversarial examples 
that introduce more noise during adversarial defending. 
Training with \textfooler examples ($\mathcal{C}_{\textfooler}$) helps, 
although most of its gain is in defending against other attacking 
methods (\hotflip and \aecf).

To summarise, these results demonstrate that our grey-box framework of 
training an augmented classifier together with a generator produces a 
more robust classifier, compared to the baseline approach of training a 
classifier using data augmented by adversarial examples.



\section{Conclusion}

In this paper, we proposed a grey-box adversarial attack and defence framework for
sentiment classification. Our framework combines a generator with two copies of the target 
classifier: a static and an updated model. Once trained, the generator 
can be used for generating adversarial
examples, while the augmented (updated) copy of the classifier is an 
improved model that is less susceptible to adversarial attacks.  Our 
results demonstrate
that the generator is capable of producing high-quality adversarial 
examples that preserve the original ground truth and is approximately an 
order of magnitude faster in creating  adversarial examples compared to 
state-of-the-art attacking methods.
Our framework of building an improved classifier together with an 
attacking generator is also shown to be more effective than the baseline 
approach of training a classifier using data augmented by adversarial 
examples. 

The combined adversarial attack and defence framework, though only evaluated
on sentiment classification, should be adapted easily to other NLP problems (except 
for the counter-fitted embeddings, which is designed for sentiment analysis). This 
framework makes it possible to train adversarial attacking models and defending models
simultaneously for NLP tasks in an adversarial manner. 



\section{Ethical Considerations}

For the human evaluation in \secref{human-evaluation}, each assignment 
was paid \$0.06 and estimated to take 30 seconds to complete, which 
gives an hourly wage of \$7.25 ($=$ US federal minimum wage).  An 
assignment refers to scoring the sentiment/coherence of a sentence, 
or scoring the semantic similarity of a pair
of sentences.


\bibliography{emnlp2020}
\bibliographystyle{acl_natbib}

\end{document}